\pdfoutput=1

\documentclass[11pt]{article}

\usepackage[final]{EACL2023}

\usepackage{times}
\usepackage{latexsym}

\usepackage{graphicx}
\usepackage{amsfonts}
\usepackage{amsmath}
\usepackage{booktabs} 

\usepackage[T1]{fontenc}

\usepackage[utf8]{inputenc}

\usepackage{microtype}

\usepackage{inconsolata}
\usepackage{microtype}

\title{Do we need Label Regularization to Fine-tune Pre-trained~Language Models?}

\author{Ivan Kobyzev$^{1}\thanks{\ \ Equal Contribution}$, Aref Jafari$^{1,2*}$, Mehdi Rezagholizadeh$^1$, Tianda Li$^1$, Alan Do-Omri$^1$,  
\\ \textbf{Peng Lu$^{1,3}$, Pascal Poupart$^2$, Ali Ghodsi$^2$}  \\
 $^1$Huawei Noah’s Ark Lab\\
$^2$ University of Waterloo, Canada \\
$^3$ Universit\'e de Montr\'eal, Canada\\
{\{ivan.kobyzev,mehdi.rezagholizadeh\}@huawei.com} \\
{\{aref.jafari, ppoupart, ali.ghodsi\}@uwaterloo.ca}
}

\begin{document}
\maketitle
\begin{abstract}
Knowledge Distillation (KD) is a prominent neural model compression technique that heavily relies on teacher network predictions to guide the training of a student model. Considering the ever-growing size of pre-trained language models (PLMs), KD is often adopted in many NLP tasks involving PLMs. However, it is evident that in KD, deploying the teacher network during training adds to the memory and computational requirements of training. In the computer vision literature, the necessity of the teacher network is put under scrutiny by showing that KD is a label regularization technique that can be replaced with lighter teacher-free variants such as the label-smoothing technique. However, to the best of our knowledge, this issue is not investigated in NLP. 
Therefore, this work concerns studying different label regularization techniques and whether we actually need them to improve the fine-tuning of smaller PLM networks on downstream tasks. In this regard, we did a  comprehensive set of experiments on different PLMs such as BERT, RoBERTa, and GPT with more than 600 distinct trials and ran each configuration five times. This investigation led to a surprising observation that KD and other label regularization techniques do not play any meaningful role over regular fine-tuning when the student model is pre-trained. 
We further explore this phenomenon in different settings of NLP and computer vision tasks and demonstrate that pre-training itself acts as a kind of regularization, and additional label regularization is unnecessary.
\end{abstract}

\section{Introduction}
\begin{figure}[tb]
  \centering
    \includegraphics[width=1\linewidth]{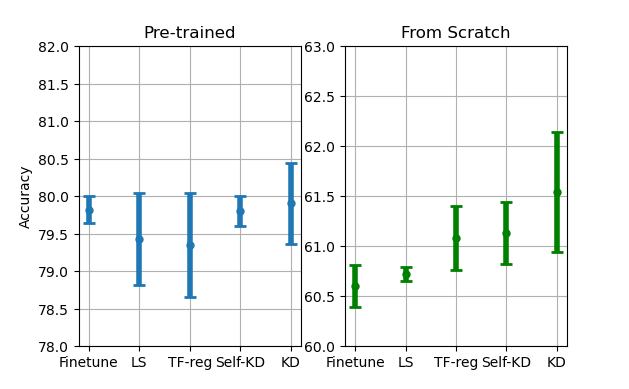} \\[\abovecaptionskip]
  \caption{DistilRoBERTa results on the test set for the average of seven GLUE tasks. Graph shows the mean performance and one standard deviation interval for the pre-trained and randomly initialized models computed over five runs. For the pre-trained model all intervals intersect, hence label regularization doesn't improve the performance,  but for the model trained from scratch label regularization methods outperform base training.}\label{fig:glue_av}
\end{figure}

Nowadays, we witness the tendency of ever-growing state-of-the-art neural networks. This is especially  more evident in natural language processing (NLP): the famous GPT-3 \cite{brown2020language} has reached 175 billion parameters and a recent Chinese pre-trained language model \cite{pangu-alpha} has 200 billion parameters. It  is shown that big over-parameterized neural networks not only have higher VC dimension, and hence more approximation ability \cite{bendavid},  but also their optimization regime is  smoother \cite{safran2020}. At the same time, the optimal point found for big networks has better generalization property \cite{Brutzkus2019WhyDL}.

One can use the advantages of trained big neural networks and  transfer their learned knowledge (weights and biases) to a smaller network. There are several approaches to such transferring techniques \cite{Cheng2017ASO}, but here we will focus on  Knowledge distillation (KD)~\cite{hinton2015distilling}, a prominent neural model compression technique,  which has been applied in many different forms across various domains \cite{gou2020knowledge}  such as computer vision and NLP. To distill knowledge from a bigger model (a teacher) to a smaller model (a student), KD adds an extra loss term to ensure the student predictions match with the teacher output.
In the NLP domain, KD is widely adopted for compressing pre-trained language models (PLMs)  \cite{sanh2019distilbert, jiao2019tinybert, jafari2021anneal}. 
The success of KD is attributed to different potential factors such as additional information presented by the dark knowledge (i.e. a term referring to the notion of class similarity information deriving from the teacher predictions which can not be found in the one-hot ground-truth labels) 
~\cite{hinton2012neural}, regularization effect of the KD loss~\cite{yuan2020}, or transferring inductive bias from one network to another~\cite{abnar2020transferring,touvron2021training}. 

\begin{figure*}[tb]
  \centering
    \includegraphics[width=\linewidth]{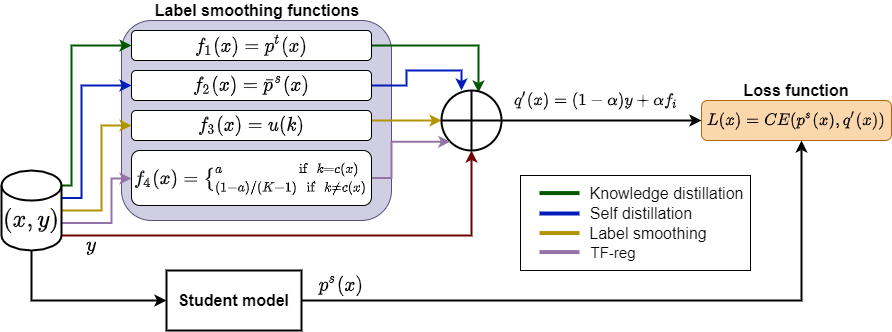} \\[\abovecaptionskip]

  \caption{All the baselines of TF/KD methods we consider can be abstracted as cross-entropy training with smoothed labels. The choice of label smoothing functions determines the method. The flow in the algorithms of each method is indicated by a colored line, where $\bigoplus$ indicates the convex combination of the one-hot label and the output of the label smoothing function. }\label{fig:vanilla}
\end{figure*}

Despite the widespread use of KD, it strongly depends on a trained teacher model, and calling the teacher during training adds to the computational cost of the training process noticeably.  
On the other hand, instead of adding the KD loss term to the student's loss, one can add a regularization term forcing the student predictions to be close to a uniform (or any arbitrary) distribution. Such a label smoothing technique results in better calibrated and more accurate classifiers \cite{muller2019}.   Recently, \citet{yuan2020} demonstrated that label smoothing can perform as well as or even outperform KD in several computer vision tasks and across various models.

This result motivated us to investigate whether the teacher-free regularization techniques (TF) can work on par or better than KD on natural language understanding tasks. In this regard, we compare KD, label smoothing, and several other teacher-free methods 
for BERT and GPT type models.  It is worth mentioning that our setting is different from the one of \citet{yuan2020}: 1) pre-trained language models are generally much bigger than the models from machine vision, and 2) classification tasks in our setting are mostly binary or three classes, compared to a hundred classes in CIFAR100 or two hundred in Tiny ImageNet. We ran the experiments multiple times to take into account the stochasticity of the training.
Overall, we show a similar pattern: teacher-free  techniques perform on par with KD methods; but we additionally observed a surprisingly different phenomenon: the gap between  base fine-tuning (without KD) and fine-tuning with label regularization (KD or TF) diminished.

We  explore the reasons why the base fine-tuning technique is a strong competitor of KD/TF regularization on NLU tasks. This situation is somewhat opposite to the one reported in computer vision. We hypothesized and tested the following potential explanations for our observations: 1) The small number of classes in GLUE datasets (usually 2 or 3 classes) in contrast to 10 or 100 classes in CV tasks; 2) Language models are extensively  pre-trained while CV models are not. Our  experiments indicate that the second hypothesis is true, whereas the number of classes doesn't play a big role in the performance gap. To the best of our knowledge, 
the effect of pre-training on fine-tuning with label regularization 
was never mentioned in the literature  and 
 deserves additional study.

Overall, our main contributions in this paper are the following:
\begin{enumerate}
    \item Thorough comparison of TF and KD methods across both BERT and GPT models on the GLUE and other NLU benchmarks (more than 600 distinct experiments overall).
    We showed that, on average, KD does not  significantly outperform the fine-tuning or TF techniques.
    \item We studied the gap between base fine-tuning and fine-tuning with KD/TF and observed that this gap is  negligible for NLU tasks. We demonstrated that this insignificant result is unlikely to be caused by the number of classes in the dataset. 
    \item We
    showed the evidence that the  pre-training  of neural networks reduces the performance gap on downstream tasks between the base training and training with label regularization both in NLP and computer vision domains. We supported this claim by performing Wilcoxon statistical test to demonstrate significance.
\end{enumerate}

\section{Background}
In this section, we give a brief overview of the KD and TF techniques we will be investigating. Everywhere in the paper, we consider a classification problem with $K$ classes. Denote by $q(x)$  the one-hot label of a data point $x$. 
 \subsection{Knowledge distillation (KD)}
 This classical method of transferring knowledge gained traction after the paper \cite{hinton2015distilling}. Assume that we have a trained network (a teacher) and a network we want to train (a student). 
  Let $p^t(x)$ and $p^s(x)$ be teacher's and student's predictions respectively. One wants to transfer the knowledge from the teacher to the student.  For that, one can formulate a total loss for KD as:
 \begin{equation}
 \label{eq:kdloss}
     L = (1-\alpha ) H(q,p) + \alpha L_{KD},
 \end{equation}
 where $H(q,p)$ is the cross-entropy loss and $L_{KD} = D_{KL}(p_\tau^t,p_\tau^s)$ is a KL divergence between the teacher's and the student's outputs scaled with the temperature $\tau$, i.e., $p_\tau(k) = softmax(z_k/\tau)$, where $z_k$ is the output logits of the model. When $\tau=1$,
 KD training is equivalent to cross-entropy training with the new labels ``smoothed" by the teacher: 
\begin{equation}
\label{eq:kd_lbls}
    q'(x) = (1-\alpha) q(x) + \alpha p^t.
\end{equation}

\subsection{Teacher-free methods}
\paragraph{Label smoothing (LS)}
As \citet{yuan2020} observed, the loss in Equation \ref{eq:kdloss} is structurally similar to the label smoothing loss, where one has to replace the term $L_{KD}$ with $L_{LS} = D_{KL}(u,p^s)$, where $u(k) = 1/K$ is the uniform distribution on $K$ classes. Training with the label smoothing loss is equivalent to cross-entropy training with smoothed labels: 
\begin{equation}
\label{eq:ls}
    q'(x) = (1-\alpha) q(x) + \alpha u .
\end{equation}
Varying the hyperparameter $\alpha$, one can change the shape of the new labels $q'$ from smoother (higher values of $\alpha$) to sharper ($\alpha$ closer to zero). 

\paragraph{TF-reg}  \citep{yuan2020}) introduced a modification of LS with a sharper label-dependent smoothing distribution. More formally, for TF-reg one switches the uniform distribution $u$ in Equation \ref{eq:ls} to a more peaky label-dependent distribution $p_c^d(k)$, defined by:
\begin{equation}
\label{eq:p_tfreg}
    p_c^d(k) = \begin{cases} a, \ \text{if} 
    \  k=c \\ (1-a)/(K-1), \ \text{otherwise.} \end{cases}
\end{equation}
The smoothed label for $x$ in TF-reg is given by:
\begin{equation}
\label{eq:tfreg}
    q'(x) = (1-\alpha) q(x) + \alpha  p_{c(x)}^d ,
\end{equation}
where $c(x)$ is the correct label for $x$. Here one has two hyperparameters ($a$ and $\alpha$) instead of just one ($\alpha$), which allows for better tuning, even though mathematically it is the same as LS. 

\citet{yuan2020} showed that LS and TF-reg perform on par or even outperform KD in machine vision for  several models and across several datasets.

\paragraph{Self distillation (Self KD)}

\citet{furlanello2018born} and  \citet{yuan2020} considered the situation where the student and the teacher have the same architectures, and a student distills the knowledge from its fine-tuned alter-ego. In particular, first, we fine-tune a copy of the student on the dataset and then freeze it. Denote its outputs by $\bar{p}^s$. Then take the second copy and train it with the cross-entropy loss with smoothed labels:
\begin{equation}
\label{eq:skd}
    q'(x) = (1-\alpha) q(x) + \alpha \bar{p}^s(x) .
\end{equation}

The summary of all the TF and KD methods we compare can be found in Figure \ref{fig:vanilla}.

\begin{table*}[h]
\centering 
\resizebox{\textwidth}{!}{
\begin{tabular}{c c c c c c c c c c} 
\hline
baseline & CoLA & RTE & MRPC & SST-2 & QNLI & MNLI & QQP & Score  \\ [0.5ex] 
\hline 
\midrule
\multicolumn{9}{c}{\textsc{Dev}}\\
\midrule
Teacher  & 68.14 & 81.23 & 91.62 & 96.44 & 94.60 & 90.23 & 91.00 & 87.67 \\
Finetune  & 60.53   \tiny{± 0.70} & 68.66 \tiny{± 1.28} & 90.58 \tiny{± 0.69} & 92.43 \tiny{± 0.16} & 90.78 \tiny{± 0.12} & 84.04 \tiny{± 0.25} & 91.44 \tiny{± 0.03} & 82.64 \tiny{± 0.11}  \\
LS   & 60.46 \tiny{± 0.74} & 69.24 \tiny{± 0.90} & 90.87 \tiny{± 0.42} & 92.75 \tiny{± 0.41} & 90.71 \tiny{± 0.09} & 83.99 \tiny{± 0.13} & 91.41 \tiny{± 0.07} & 82.78 \tiny{± 0.15}  \\
TF-reg & 60.74 \tiny{± 0.98} & 68.81 \tiny{± 0.98} & 90.78 \tiny{± 0.77} & 92.68 \tiny{± 0.15} & 91.13 \tiny{± 0.43} & 83.86 \tiny{± 0.17} & 91.45 \tiny{± 0.10} & 82.78 \tiny{± 0.29}  \\
Self-KD  & 60.48 \tiny{± 0.59} & 69.24 \tiny{± 1.20} & 90.97 \tiny{± 0.46} & 92.43 \tiny{± 0.29} & 90.91 \tiny{± 0.27} & 84.00 \tiny{± 0.19} & 91.62 \tiny{± 0.09} & 82.81 \tiny{± 0.19} \\
KD  & 62.13 \tiny{± 0.67} & 68.66 \tiny{± 1.24} & 90.83 \tiny{± 0.31} & 92.73 \tiny{± 0.34} & 91.23 \tiny{± 0.29} & 84.34 \tiny{± 0.22} & 91.68 \tiny{± 0.08} & 83.08 \tiny{± 0.14} \\
\midrule
\multicolumn{9}{c}{\textsc{Test}}\\
\midrule
Teacher  & 65.1 & 82.6 & 89.5 & 92.1 & 91.5 & 84.3 & 88.7 & 84.82 \\
Finetune  & 51.62 \tiny{± 0.96} & 62.70 \tiny{± 0.41} & 88.12 \tiny{± 0.35} & 93.22 \tiny{± 0.46} & 90.66 \tiny{± 0.15} & 83.52 \tiny{± 0.26} & 88.92 \tiny{± 0.19} & 79.82 \tiny{± 0.18} \\
LS  & 49.46 \tiny{± 3.84} & 62.52 \tiny{± 0.38} & 87.94 \tiny{± 0.51} & 93.42 \tiny{± 0.33} & 90.26 \tiny{± 0.43} & 83.34 \tiny{± 0.32} & 89.04 \tiny{± 0.08} & 79.43 \tiny{± 0.62} \\ 
TF-reg  & 49.16 \tiny{± 3.82} & 62.92 \tiny{± 0.32} & 87.44 \tiny{± 0.74} & 93.26 \tiny{± 0.28} & 90.28 \tiny{± 0.21} & 83.36 \tiny{± 0.36} & 89.04 \tiny{± 0.08} & 79.35 \tiny{± 0.70} \\
Self-KD & 51.56 \tiny{± 1.04} & 62.88 \tiny{± 0.82} & 87.92 \tiny{± 0.43}  & 93.10 \tiny{± 0.11} & 90.58 \tiny{± 0.27} & 83.46 \tiny{± 0.29} & 89.12 \tiny{± 0.10} & 79.80 \tiny{± 0.20} \\
KD   & 50.28 \tiny{± 3.07} & 63.04 \tiny{± 0.43} & 88.80 \tiny{± 0.54} & 93.44 \tiny{± 0.48} & 90.74 \tiny{± 0.26} & 83.64 \tiny{± 0.21} & 89.42 \tiny{± 0.04} & 79.91 \tiny{± 0.54} \\
\hline 
\end{tabular}
}
\caption{DistilRoBERTa results on the dev and test sets for the GLUE benchmark. F1 scores are reported for MRPC, Matthew's Correlation for CoLA, and accuracy scores for all other tasks. The teacher is RoBERTa-large. Averages and standard deviations are over 5 runs.} 
\label{t:distilroberta} 
\end{table*}

\begin{table*}[hbt!]
\centering 
\resizebox{\textwidth}{!}{
\begin{tabular}{c c c c c c c c c c} 
\hline
baseline & CoLA & RTE & MRPC & SST-2 & QNLI & MNLI & QQP & Score  \\ [0.5ex] 
\hline 
\midrule
\multicolumn{9}{c}{\textsc{Dev}}\\
\midrule
Teacher  & 65.80 & 71.48  & 89.38 & 92.77 & 92.82 & 86.3 & 91.45 & 82.19 \\
Finetune  & 41.76 \tiny{± 1.09} & 65.13 \tiny{± 1.22} & 87.09 \tiny{± 0.62} & 88.83 \tiny{± 0.27} & 86.96 \tiny{± 0.11} & 78.46 \tiny{± 0.13} & 90.02 \tiny{± 0.08} & 76.89 \tiny{± 0.25} \\
LS   & 41.97 \tiny{± 1.63} & 65.85 \tiny{± 1.16} & 87.41 \tiny{± 0.55} & 88.56 \tiny{± 0.24} & 86.90 \tiny{± 0.16} & 78.51 \tiny{± 0.14}  & 90.04 \tiny{± 0.03} & 77.03 \tiny{± 0.24} \\
TF-reg   & 42.13 \tiny{± 0.74} & 64.98 \tiny{± 1.57} & 87.19 \tiny{± 0.43} & 88.58 \tiny{± 0.31} & 86.96 \tiny{± 0.13} & 78.54 \tiny{± 0.11} & 90.02 \tiny{± 0.04} & 76.91 \tiny{± 0.18} \\
Self-KD & 41.52 \tiny{± 1.74} & 65.63 \tiny{± 1.52} & 86.73 \tiny{± 0.25} & 88.72 \tiny{± 0.28} & 86.74 \tiny{± 0.58} & 78.63 \tiny{± 0.29} & 90.08 \tiny{± 0.09} & 76.86 \tiny{± 0.39} \\
KD  & 42.48 \tiny{± 1.34} & 65.42 \tiny{± 0.95} & 88.56 \tiny{± 0.40} & 88.60 \tiny{± 0.60} & 87.31 \tiny{± 0.22} & 78.73 \tiny{± 0.19} & 90.23 \tiny{± 0.08} & 77.33 \tiny{± 0.15} \\
 \midrule
\multicolumn{9}{c}{\textsc{Test}}\\
\midrule
Teacher  & 63.8 & 69.2 & 85.1 & 89.7 & 89.2 & 83.2  & 86.2 & 80.91 \\
Finetune    & 38.58 \tiny{± 0.87} & 62.74 \tiny{± 0.31} & 83.12 \tiny{± 0.42} & 89.56 \tiny{± 0.65} & 86.62 \tiny{± 0.65} & 78.26 \tiny{± 0.27} & 87.80 \tiny{± 0.17} & 75.24 \tiny{± 0.29} \\
 LS & 40.08 \tiny{± 0.58} & 62.84 \tiny{± 0.22} & 83.24 \tiny{± 0.56} & 89.88 \tiny{± 0.50} & 86.60 \tiny{± 0.79} & 78.48 \tiny{± 0.26} & 87.78 \tiny{± 0.15} & 75.56 \tiny{± 0.14} \\
 TF-reg      & 38.92 \tiny{± 1.25} & 60.70 \tiny{± 3.03} & 82.92 \tiny{± 0.50} & 89.82 \tiny{± 0.42} & 86.22 \tiny{± 0.70} & 78.16 \tiny{± 0.33} & 87.78 \tiny{± 0.12} & 74.93 \tiny{± 0.63} \\
Self-KD & 38.92 \tiny{± 2.44} & 61.32 \tiny{± 1.26} & 83.12 \tiny{± 0.64} & 89.82 \tiny{± 0.43} & 86.60 \tiny{± 0.30} & 78.22 \tiny{± 0.53} & 87.76 \tiny{± 0.05} & 75.11 \tiny{± 0.41} \\
KD    & 38.26 \tiny{± 2.20} & 62.32 \tiny{± 1.04} & 84.74 \tiny{± 0.65} & 89.96 \tiny{± 0.22} & 86.58 \tiny{± 0.27} & 78.34 \tiny{± 0.20} & 88.02 \tiny{± 0.12} & 75.46 \tiny{± 0.51} \\
\hline 
\end{tabular}
}
\caption{BERT-small results on the dev and test sets of the GLUE benchmark. F1 scores are reported for MRPC, Matthew's Correlation for CoLA, and accuracy scores for all other tasks. The teacher is BERT-large. Averages and standard deviations are over 5 runs.} 
\label{t:bertsmall} 
\end{table*}

\section{Experiments on GLUE benchmark}
Inspired by the results of \citet{yuan2020} in machine vision, we wanted to investigate the performance of TF training on NLP data.
In this section, we evaluate the performance of the methods introduced in the Background section.

\subsection{Dataset}
We considered seven classification datasets of
the General Language Understanding Evaluation
(GLUE) benchmark \cite{wang2018glue}. 
These datasets
include linguistic acceptability (CoLA), sentiment analysis (SST-2),
paraphrasing (MRPC and QQP), Natural Language Inference (MNLI, RTE) and
Question Answering (QNLI). Notice that unlike most of the popular datasets in computer vision, GLUE tasks are either binary or ternary classification (only  MNLI has three classes).

\subsection{Experimental Setup}
We explored all the KD/TF methods in three different setups to check the consistency of the results across different models.
Our first student is DistilRoBERTa \cite{sanh2019distilbert}. It has 6 layers, 768 hidden dimensions, 8 attention heads, and 82 million parameters. In the KD scenarios, we use RoBERTa-large \cite{liu2019roberta} as its teacher (it has 24 layers, 1024 hidden dimensions, 16 attention heads, and 355 million parameters).
In the second experiment,  we use the BERT-small \cite{turc2019wellread} model with 4 layers, 512 hidden dimensions, 8 heads, and 28.7 million parameters. As a teacher, we use BERT-large \cite{devlin2018bert}
with 24 layers, 1024 hidden dimensions, and 336 million parameters.
The third student is DistilGPT-2 with 6 layers, 768 hidden dimensions, and 82 million parameters. As its teacher, the 12-layer GPT-2  \cite{radford2019language} model is used with the 768 hidden dimensions and 117 million parameters. For all these setups, we use the pre-trained models from Huggingface \cite{Wolf2019HuggingFacesTS}. All the hyperparameters and the process of their tuning  are reported in the Appendix in more detail.

\paragraph{Hardware Setup} For our experiments, we used 8 NVIDIA TESLA V100 GPUs. Each task is trained on a single GPU.

\begin{table*}[h]
\centering 
\resizebox{\textwidth}{!}{
\begin{tabular}{c c c c c c c c c c} 
\hline
baseline & CoLA & RTE & MRPC & SST-2 & QNLI & MNLI & QQP & Score  \\ [0.5ex] 
\hline 
\midrule
\multicolumn{9}{c}{\textsc{Dev}}\\
\midrule
Teacher  & 43.2 & 66.8 & 87.6 & 92.2 & 88.6 & 82.3 & 89.5  & 78.6  \\
Finetune & 38.20 \tiny{± 1.23}  & 64.92 \tiny{± 0.92} & 87.74  \tiny{± 0.34} & 91.54  \tiny{± 0.34} & 86.48  \tiny{± 0.52} & 79.93  \tiny{±0.08} & 89.70  \tiny{± 0.06} & 77.13  \tiny{±0.34} \\
 LS &  38.24  \tiny{± 1.25} & 64.84  \tiny{± 0.66} & 87.50  \tiny{± 0.28} & 91.54  \tiny{± 0.25} & 86.56 \tiny{± 0.36} & 80.14  \tiny{± 0.14} & 89.67  \tiny{± 0.10}  & 76.93  \tiny{± 0.27} \\
TF-reg & 38.04  \tiny{± 1.23}  & 64.90  \tiny{± 0.62} & 87.58  \tiny{± 0.26} & 91.34  \tiny{± 0.14} & 86.72 \tiny{± 0.34}  & 80.14  \tiny{± 0.22}  & 89.64  \tiny{± 0.08} & 76.91  \tiny{± 0.27}  \\
Self-KD &  39.41  \tiny{± 0.91} & 65.62  \tiny{± 1.61} & 87.24  \tiny{± 0.21} & 90.84  \tiny{± 0.31} & 87.04  \tiny{±  0.17} & 80.57  \tiny{± 0.16} & 89.83 \tiny{± 0.04} & 77.16  \tiny{± 0.33} \\ 
KD  & 38.94  \tiny{± 1.10} & 66.80  \tiny{± 0.82} & 87.22  \tiny{± 0.74} & 90.86  \tiny{± 0.33} & 86.82  \tiny{± 0.32} & 80.30  \tiny{± 0.17} & 89.97  \tiny{± 0.24} & 77.34  \tiny{± 0.30} \\
\midrule
\multicolumn{9}{c}{\textsc{TEST}}\\
\midrule
Teacher  & 46.7 & 65.0 & 86.4 & 88.3 & 88.5 & 81.8 & 87.9 & 77.8 \\
Finetune  & 31.00  \tiny{± 1.32} & 60.52  \tiny{± 0.66} & 84.52  \tiny{± 0.56} & 90.22  \tiny{± 1.08} & 85.34  \tiny{± 0.30} & 79.84  \tiny{± 0.22} & 87.68  \tiny{± 0.12} & 74.16  \tiny{± 0.28}  \\
 LS & 31.30  \tiny{± 2.00} & 60.18  \tiny{± 0.63} & 84.68  \tiny{± 0.41} & 91.18  \tiny{± 0.41} & 85.28  \tiny{± 0.30} & 79.78 \tiny{± 0.19}  & 87.76 \tiny{± 0.14} & 74.31  \tiny{± 0.25} \\
TF-reg & 31.74  \tiny{± 2.06} & 60.22  \tiny{± 0.40} & 84.50  \tiny{± 0.46} & 90.62  \tiny{± 0.56} & 85.38  \tiny{± 0.32} & 79.68  \tiny{± 0.28} & 87.24 \tiny{± 0.50} & 74.20  \tiny{± 0.40} \\
Self-KD & 35.28  \tiny{± 1.55} & 61.02  \tiny{± 1.23} & 83.72  \tiny{± 0.50} & 90.30  \tiny{± 0.54} & 86.14  \tiny{± 0.33} & 80.12  \tiny{± 0.15} & 87.86  \tiny{± 0.16} & 74.92  \tiny{± 0.18} \\
KD  & 32.96 \tiny{± 2.84} & 60.40  \tiny{± 0.20} & 84.76  \tiny{± 0.56} & 90.38  \tiny{± 0.53} & 85.82  \tiny{± 0.15} & 80.10  \tiny{± 0.14} & 88.08 \tiny{± 0.14} & 74.64 \tiny{± 0.43} \\
\hline 
\end{tabular}
}
\caption{DistilGPT-2 results on the dev and test sets of the GLUE benchmark. F1 scores are reported for MRPC, Matthew's Correlation for CoLA, and accuracy scores for all other tasks. The teacher is GPT-2 (12 layers). Averages and standard deviations are over 5 runs.} 
\label{t:distilgpt} 
\end{table*}

\subsection{Results}
\paragraph{DistilRoBERTa} We start with conducting the GLUE experiments over the  DistilRoBERTa model. We report the results on GLUE dev and test sets in Table \ref{t:distilroberta}. On the dev set, we observe the following patterns: 1) The teacher-free methods (LS, TF-reg, Self-KD) outperform the Finetune baseline; 2) KD is the best technique but the standard deviation intervals intersect with the TF baselines.

Although the results of the dev set in the first experiments follow the trends of TF results in CV, examining the test results reveals some irregularities (Table~\ref{t:distilroberta}).    
 In particular, we  observe that: 1) all the TF regularization techniques perform slightly worse than  Finetune; 2) KD is on average the best technique, but it is comparable with Finetune up to one standard deviation. See Figure~\ref{fig:glue_av} (left) for the summary.

\paragraph{BERT-small } In the second experiment, we evaluate the BERT-small model. Even here the story is more or less similar to our first experiment on DistilRoBERTa. Results are reported in Table~\ref{t:bertsmall}.  On the dev set, we observe that: 1) TF performs on par (up to one standard deviation) with Finetune while LS is slightly better. 2) KD is the best performer, but standard deviation intervals intersect with some of the TF baselines.
 On the test set,  we observe that all the methods perform more or less on par up to one standard deviation while LS is slightly better. 

\paragraph{DistilGPT-2} For DistilGPT-2, we see roughly similar patterns as in our previous experiments (Table \ref{t:distilgpt}). On both dev and test sets, all the methods' performance is more or less similar, with the standard deviation intervals overlapping.

The overall conclusion of our experiments  is that, on average, KD or TF methods are slightly better, but the gap between the regularization techniques and the pure fine-tuning technique is not significant. Our results on the GLUE benchmark are very different from the reported results in the CV domain  where pure fine-tuning without TF or KD underperforms.  
To explain the results, 
we formulate some hypotheses and scrutinize them with more experiments in the next section. 

\section{Analysis}
In this section, we investigate potential reasons for getting the negligible difference in the relative performance of base fine-tuning, teacher-free training, and KD. We conduct some experiments to evaluate two particular hypotheses we have as potential reasons behind these inconsistencies: 1) the number of classes in the GLUE tasks is much lower than for the CV tasks; 2) NLU models are pre-trained and pre-training can attenuate the regularization impact of KD and TF methods. In the remainder of this section, we will go over some new experiments which were done to evaluate these two hypotheses respectively.   
\subsection{Hypothesis 1: Number of Classes}
\subsubsection{SST-5}
SST-5 is a fine-grained sentiment classification dataset with 5 classes introduced in  \cite{socher2013recursive}. 
 We consider a setting of DistilRoBERTa student and RoBERTa-large (24 layers) teacher. 
 We ran experiments for 5 seeds. The results are presented in Table \ref{t:sst5}. Overall, we can see that the standard deviations of the results are quite big, which prevents us from concluding that any technique is superior. Similar to GLUE, we note that the gap between fine-tuning and TF/KD is not observed. 

In our next experiment, we increase  the number of classes even more to see if the gap appears.

\begin{table}[hbt!]
\centering 
\begin{tabular}{c c c} 
\hline
 baseline    &  Accuracy (dev) & Accuracy (test) \\
\hline
Teacher    & 56.86   & 59.95  \\
 Finetune    & 53.40 \tiny{± 0.85}  & 54.43 \tiny{± 0.56}  \\
LS    & 53.50 \tiny{± 0.98}  &  53.96 \tiny{± 0.77}  \\
 TF-reg   &  53.62 \tiny{± 0.90}    &  53.93 \tiny{± 0.42} \\ 
 KD  & 53.59 \tiny{± 0.26}  &  54.14 \tiny{± 0.92}  \\
\hline 
\end{tabular}
\caption{DistilRoBERTa results on SST-5. Averages and standard deviations are over 5 runs.} 
\label{t:sst5} 
\end{table}

\subsubsection{FewRel}
\citet{han-etal-2018-fewrel} introduced this dataset for relation classification. Originally, this dataset was designed for few-shot learning, so we had to slightly modify it for our purpose. First, we consider the train set of FewRel. It has 64 classes and each class has 700 instances. We shuffle the data for each class and allocated 500 instances to our train set, 100 to our dev set, and 100 to our test set. We perform the experiments five times and get a new dataset for each seed, as recommended by \citet{bouthillier2021variance}.
The detailed procedure can be found in Appendix. Overall, we generated  a text classification dataset with 64 classes. 

We took DistilRoBERTa as a student and RoBERTa-base (12 layers) as a teacher. We ran experiments for 5 seeds and tuned hyperparameters  for the first one (see Appendix for details). The results are in Table~\ref{t:fewrel}. We can observe that all the methods perform similarly up to one standard deviation and we don't see a gap between Finetune and KD/TF again. 

As a conclusion of SST-5 (5 classes) and FewRel (64 classes) experiments,  we do not see  any evidence that the number of classes in classification tasks affects the gap.

\begin{table}[hbt!]
\centering 
\begin{tabular}{c c c} 
\hline
 baseline    &  Accuracy (dev) & Accuracy (test) \\
\hline
Teacher    &  88.93 \tiny{± 0.27}  &  88.63 \tiny{± 0.45}  \\
 Finetune    & 86.31 \tiny{± 0.32} & 86.28 \tiny{± 0.51}  \\
LS    & 86.35 \tiny{± 0.31}  & 86.22 \tiny{± 0.47}  \\
 TF-reg   &  86.35 \tiny{± 0.31}   & 86.26 \tiny{± 0.47}  \\ 
 KD  & 86.66 \tiny{± 0.36} & 86.41 \tiny{± 0.54}   \\
\hline 
\end{tabular}
\caption{DistilRoBERTa results on FewRel (64 classes). Averages and standard deviations are over 5 runs.} 
\label{t:fewrel} 
\end{table}

\subsection{Hypothesis 2: Effect of Pre-training}
We pose the following question: what is the major difference between Language Models and Computer Vision Models that might affect the performance gap between base training and label regularization? As an immediate hypothesis, we thought that extensive pre-training of the models we experimented with in the previous sections might be the reason.

\begin{table}[hbt!]
\centering 
\begin{tabular}{c c c} 
\hline
 baseline & From scratch & Pre-trained \\
\hline
 Base & 77.04 \tiny{± 0.26}  & 78.17 \tiny{± 0.27}  \\
 LS  & 78.01 \tiny{± 0.20} &  78.67	\tiny{± 0.20}  \\
 TF-reg & 78.16 \tiny{± 0.20}  & 78.94 \tiny{± 0.28} \\
\hline 
\end{tabular}
\caption{ResNet18 on CIFAR100. Pre-training is done on the ImageNet dataset. Averages and standard deviations are over 10 runs.} 
\label{t:resnet} 
\end{table}

\begin{table*}[h]
\centering 
\begin{tabular}{c c c c c c c c c c} 
\hline
baseline & CoLA & RTE & MRPC & SST-2 & QNLI & MNLI & QQP & Score  \\ [0.5ex] 
\hline 
\midrule
\multicolumn{9}{c}{\textsc{Dev}}\\
\midrule
Teacher  & 68.14 & 81.23 & 91.62 & 96.44 & 94.60 & 90.23 & 91.00 & 87.67 \\
Base  & 13.3   \tiny{± 0.9} & 53.0 \tiny{± 0.4} & 81.4 \tiny{± 0.3} & 81.2 \tiny{± 0.5} & 60.8 \tiny{± 0.6} & 62.1 \tiny{± 1.1} & 80.8 \tiny{± 0.2} & 61.87 \tiny{± 0.21}  \\
LS   & 14.0 \tiny{± 0.9} & 53.8 \tiny{± 1.0} & 82.1 \tiny{± 0.7} & 81.7 \tiny{± 0.5} & 61.5 \tiny{± 0.6} & 62.8 \tiny{± 0.8} & 81.1 \tiny{± 0.5} & 62.45 \tiny{± 0.45}  \\
TF-reg & 14.4 \tiny{± 1.2} & 53.0 \tiny{± 0.5} & 82.3\tiny{± 0.4} & 81.8 \tiny{± 0.4} & 61.5 \tiny{± 1.3} & 62.7 \tiny{± 1.1} & 80.6 \tiny{± 0.2} & 62.35 \tiny{± 0.25}  \\
Self-KD  & 14.3 \tiny{± 0.5} & 53.0 \tiny{± 0.4} & 82.3 \tiny{± 0.5} & 81.4 \tiny{± 0.1} & 61.3 \tiny{± 0.4} & 63.5 \tiny{± 1.2} & 80.7 \tiny{± 0.1} & 62.39 \tiny{± 0.28} \\
KD  & 16.6 \tiny{± 0.7} & 53.6\tiny{± 0.9} & 81.8 \tiny{± 0.4} & 81.2\tiny{± 0.5} & 61.4 \tiny{± 0.4} & 63.1 \tiny{± 0.3} & 81.6 \tiny{± 0.1} & 62.80 \tiny{± 0.45} \\
\midrule
\multicolumn{9}{c}{\textsc{Test}}\\
\midrule
Teacher  & 65.1 & 82.6 & 89.5 & 92.1 & 91.5 & 84.3 & 88.7 & 84.82 \\
Base  & 9.8 \tiny{± 0.1} & 51.6 \tiny{± 1.3} & 79.6 \tiny{± 0.3} & 80.3 \tiny{± 0.0} & 60.7 \tiny{± 0.5} & 61.4 \tiny{± 0.3} & 80.8\tiny{± 0.2} & 60.60 \tiny{± 0.21} \\

LS  & 10.5 \tiny{± 0.2} & 53.0 \tiny{± 0.0} & 80.0 \tiny{± 0.4} & 80.9 \tiny{± 0.4} & 60.7 \tiny{± 0.7} & 61.7 \tiny{± 0.1} & 81.1 \tiny{± 0.7} & 60.72 \tiny{± 0.07} \\ 

TF-reg  & 11.8 \tiny{± 1.1} & 51.8 \tiny{± 1.5} & 79.2 \tiny{± 1.5} & 81.2 \tiny{± 0.3} & 60.9 \tiny{± 0.4} & 61.7 \tiny{± 0.1} & 81.2 \tiny{± 0.4} & 61.08 \tiny{± 0.32} \\

Self-KD & 11.5 \tiny{± 0.3} & 51.0 \tiny{± 0.7} & 79.5 \tiny{± 1.2}  & 80.6 \tiny{± 0.1} & 61.9 \tiny{± 0.4} & 62.3 \tiny{± 0.2} & 81.3 \tiny{± 0.1} & 61.13 \tiny{± 0.31} \\

KD   & 12.8 \tiny{± 2.8} & 51.5 \tiny{± 1.1} & 79.3 \tiny{± 0.6} & 82.1 \tiny{± 0.4} & 61.2 \tiny{± 0.2} & 62.6 \tiny{± 0.0} & 81.4 \tiny{± 0.4} & 61.54 \tiny{± 0.60} \\

\hline 
\end{tabular}
\caption{Randomly initialized DistilRoBERTa results on the dev and test sets for the GLUE benchmark. F1 scores are reported for MRPC, Matthew's Correlation for CoLA, and accuracy scores for all other tasks. The teacher is RoBERTa-large. Averages and standard deviations are over 5 runs.} 
\label{t:distilroberta_random} 
\end{table*}

\subsubsection{Computer vision} 
First, we did a sanity check and performed some experiments from \cite{yuan2020} for multiple seeds. In the paper, they didn't mention the standard deviation, but it is important for us to check if the gap we hoped to find is not a result of randomness. We considered CIFAR100 \cite{Krizhevsky2009Learning} and trained ResNet18 student \cite{He2016DeepRL} without label regularization and with LS and TF-reg techniques. 
At the same time we repeated similar experiments, but now with ResNet18 pre-trained on ImageNet dataset \cite{russakovsky2015imagenet}.
The results are reported in Table \ref{t:resnet}. 
Here, we can see that for the unpretrained model the standard deviation intervals between base training and label regularization don't intersect and the gap is reasonably large. However, the gap diminishes notably for the pre-trained model. 

This gives us some initial evidence that our hypothesis might be true.  In our next experiment, we report results that support this hypothesis.

\subsubsection{NLU}
\paragraph{GLUE experiments}
To investigate the effect of pre-training on the relative performance, we took a model with the same architecture as DistilRoBERTa, but instead of initializing it with pre-trained weights, we randomly initialize it (with normal distribution using the built-in Huggingface function). We used the hyperparameters from the pre-trained experiments. Then we ran experiments for 5 seeds. The results are reported in Table~\ref{t:distilroberta_random}. 
We can see that, unlike the pre-trained model, the gap between base training and all the label regularization methods is bigger and the intersection of standard deviation intervals is much smaller or nonexistent. See Figure~\ref{fig:glue_av} (right) for the summary.

\paragraph{SST-5 experiments}
As a next step, we wanted to formally check the statistical significance of the findings that we reported in the previous sections. For this, we considered again the SST-5 dataset and trained both the pre-trained and randomly initialized DistilRoBERTa on it. We aim to determine whether there is a statistically significant difference between base training and TF/KD training for each of the pre-trained and randomly initialized cases. We used the (two-sided) Wilcoxon signed-rank test \cite{wilcoxon1945individual} over the results of eight random seeds. The Wilcoxon test is a non-parametric statistical test that checks the null hypothesis, i.e., whether two related paired samples come from the same distribution. The results are reported in Table~\ref{t:wilcoxon_sst5}. We can see that for the pre-trained model there is no statistically significant difference between base training and label regularization (p-value is greater than 0.05). However, if the model is trained from scratch, the difference becomes statistically significant. 

We also tried state-of-the-art KD method, Annealing KD (AKD) \cite{jafari2021anneal} which is  like vanilla KD doesn't require data augmentation or an access to teacher's intermediate layers. The result of the Wilcoxon test (Table~\ref{t:wilcoxon_sst5}) shows that similarly it doesn't give a significantly better performance for a pre-trained model.



\begin{table}[hbt!]
\centering 
\begin{tabular}{c c c} 
\hline
 Comparison & Pre-trained & From scratch \\
\hline
 Base vs TF-reg & 0.46  & 0.01  \\
 Base vs KD  & 0.94  & 0.01 \\
 Finetune vs AKD & 0.74 & - \\
\hline 
\end{tabular}
\caption{Wilcoxon signed-rank test results for DistilRoBERTa model trained on SST-5 dataset. P-values of the test are reported with p-value less than 0.05 meaning the difference is significant. The results are over the test results of 8 runs.} 
\label{t:wilcoxon_sst5} 
\end{table}


\section{Related Work}
Our finding that pre-training reduces or even removes the gap between base training and TF/KD training can serve as an indication of a regularization property of pre-training. Several works are exploring this in the literature.  

\citet{Tu2020} studied the relation between pre-training and spurious correlations. They demonstrated that pre-trained models are more robust to spurious correlations because they can generalize from a minority of training examples that counter the spurious pattern. 
 \citet{Furrer2020} demonstrated that Masked Language Model pre-training helps in semantic parsing scenarios to improve compositional generalization. The authors hypothesize that the primary benefit provided by MLM pre-training is the improvement of the model’s ability to substitute similar words or word phrases by ensuring they are close to each other in the representation space. 
 \citet{turc2019wellread} showed that pre-training is very beneficial for smaller architectures, and fine-tuning pre-trained compact models can be competitive with more elaborate methods.

\section{Discussion and Future Work}
We started this comparison of KD and TF regularizations on NLU tasks in the hope that a pattern similar to the one in computer vision \cite{yuan2020} will emerge. In particular, we expected to see TF and KD  perform on par while outperforming Finetune. However, it turned out that the gap, even if it exists for some seeds, is not significant.  

We further scrutinized the gap between Finetune and KD/TF regularization.
We hypothesized that the lack of this gap in NLU might be the result of a small number of classes in GLUE classification tasks, however, this doesn't seem to be the case: experiments on SST-5 (5 classes) and FewRel (64 classes) datasets didn't show a significant gap either.
We showed that another hypothesis is likely to be true: the extensive pre-training of Language Models  erases the gap. The application of a statistical test confirms that a non-negligible gap appears when models are trained from scratch. 
It seems that pre-training discovers a good enough initialization for fine-tuning so that even basic unregularized training can find a solution as good as training with (TF or KD) regularization. A rigorous explanation of this phenomenon is an interesting challenge for future work. 

We would like to add one important remark. Our finding of this work does not suggest disregarding KD or other types of regularizations in NLP but rather using the more advanced or enhanced versions of these techniques. First of all, as shown in several works in the literature~\cite{sanh2019distilbert, sun2019patient, turc2019wellread, jiao2019tinybert, tahaei2021kroneckerbert}, KD is very important for the pre-training stage of the student models. 
Similarly, \citet{gao-etal-2020-towards} demonstrated the value of label smoothing for training machine translation models.

Moreover, improved variants of KD might still facilitate fine-tuning of pre-trained models. 
Even when vanilla KD doesn't give a statistically significant advantage over base fine-tuning, several works in the literature show that improved versions of KD with different auxiliary training schemes could be beneficial. For example, one can incorporate intermediate layer distillation \cite{sun2019patient, passban2021alpkd,wu2020skip, wu2021universal}, data augmentation \cite{rashid2021matekd, kamalloo2021not} or contrastive training~\cite{sun2020contrastive}.  
Investigating better KD techniques or, more generally, better regularization methods that can improve the fine-tuning of PLMs even further will be an important direction for future work.

\section*{Limitations}
In the current work we present an extensive empirical evidence that label regularization doesn't improve fine-tuning of a pre-trained model. However, we don't have any theoretical explanation of this puzzling phenomenon. Understanding the interactions of different regularization methods and how they affect the optimization is a highly nontrivial problem.

\section*{Acknowledgments}
We thank Mindspore,\footnote{\url{https://www.mindspore.cn/}} which is a new deep learning computing framework, for partial support of this work.




\bibliographystyle{acl_natbib}
\bibliography{custom}

\appendix

\section{Hyperparameters for GLUE and SST-5 experiments}
We ran experiments for seeds 42,	549, 1237, 230 and 805. 
For all baselines we run experiments for 30 epoch.
All hyperparameters for pre-trained and randomly initialized models are listed in the Tables~\ref{table:hyper_roberta}, \ref{table:hyper_gpt}.

When we do 8 seeds for the statistical test, we add seeds 4653,	5589 and	992.

Hyperparameters for Annealing KD on SST-5 are listed in Table~\ref{table:AKD}

\begin{table*}[hbt!]
\centering 
\begin{tabular}{c c } 
\hline 
Hyper-parameter & Value \\  
\hline 
Learning rate & $2\cdot 10^{-5}$  \\
Batch Size & 32 \\ 
Temperature & 1  \\
Training epoch & 30  \\
$\alpha$ for LS & 0.1 \\
$\alpha$ for KD, Self-KD and TF-reg  & 0.5 \\
$\alpha$ for KD and TF-reg RI SST-5  & 0.9 \\
$a$ for TF-reg & 0.95 \\
\hline 
\end{tabular}
\caption{Hyperparameters for DistilRoBERTa and BERT-Small models on GLUE and SST-5 for pre-trained and randomly initialized (RI) models} 
\label{table:hyper_roberta} 
\end{table*}

\begin{table*}[hbt!]
\centering 
\begin{tabular}{c c c c c c c c} 
\hline 
Hyper-parameter & CoLA & RTE & MRPC &  SST-2 & QNLI & QQP & MNLI \\  
\hline 
Learning rate & $10^{-5}$ & $2\cdot 10^{-5}$ & $10^{-5}$ & $2\cdot 10^{-5}$ & $10^{-5}$ & $2\cdot 10^{-5}$ & $2\cdot 10^{-5}$ \\
Batch Size & 16 & 16 & 16 & 16 & 16 & 16 & 16 \\ 
Temperature & 1 & 1 & 1 & 1 & 1 & 1 & 1 \\
Training epoch & 30 & 30 & 30 & 30 & 30 & 30 & 30 \\
$\alpha$ for LS & 0.1 & 0.1 & 0.1 & 0.1 & 0.1 & 0.1 & 0.1 \\
$\alpha$ for KD & 0.5 & 0.5 & 0.5 & 0.5 & 0.5 & 0.5 & 0.5 \\
$\alpha$ for Self-KD & 0.5 & 0.5 & 0.5 & 0.5 & 0.5 & 0.5 & 0.5 \\
$\alpha$ for TF-reg & 0.5 & 0.5 & 0.5 & 0.5 & 0.5 & 0.5 & 0.5 \\
$a$ for TF-reg & 0.95 & 0.95 & 0.95 & 0.95 & 0.95 & 0.95 & 0.95 \\
\hline 
\end{tabular}
\caption{Hyperparameters for DistilGPT-2 on GLUE tasks for Finetune and regular KD and TF} 
\label{table:hyper_gpt} 
\end{table*}

\begin{table*}[hbt!]
\centering 
\begin{tabular}{c c } 
\hline 
Hyper-parameter & Value \\  
\hline 
Learning rate & $2\cdot 10^{-5}$  \\
Batch Size & 8 \\ 
Max Temperature & 10  \\
Training epochs Phase I & 20  \\
Training epochs Phase II & 10 \\
\hline 
\end{tabular}
\caption{Hyperparameters for Annealing KD for DistilRoBERTa on SST-5} 
\label{table:AKD} 
\end{table*}

\section{Experiments on FewRel}
\subsection{How we constructed the dataset}
For each seed (42,	549, 1237, 230 and 805) separately we constructed a new dataset. 

Train set of FewRel has 64 classes, each class has 700 instances. We shuffle the data for each class (with a current seed) and allocated first 500 instances for our train set, second 100 for our dev set and last 100 for our test set. 

We concatenated the  context, head, and tail of the relation into one piece of text to be classified. 

\subsection{Hyperparameters}
All hyperparameters are shown in Table~\ref{table:hyper_fewrel}
\begin{table}[hbt!]
\centering 
\begin{tabular}{c c} 
\hline 
Hyper-parameter & Value \\  
\hline 
Learning rate & $1.5 \cdot 10^{-5}$ \\
Batch Size & 32 \\ 
Temperature & 1 \\
Training epoch & 30  \\
$\alpha$ for LS & 0.1 \\
$\alpha$ for KD & 0.5\\
$\alpha$ for TF-reg & 0.5 \\
$a$ for TF-reg & 0.95 \\
\hline 
\end{tabular}
\caption{Hyperparameters for DistilRoBERTa on FewRel for Finetune, KD and TF}
\label{table:hyper_fewrel} 
\end{table}

\section{Experiments on CIFAR100}

We follow the experimental setup~\cite{yuan2020}.
For optimization we used SGD with a momentum of 0.9. The learning rate starts at 0.1 and is then divided by 5 at epochs 60, 120 and 160. All experiments are repeated 10 times with different random initialization. The seeds we used: 11, 125, 1350, 23, 230, 4653, 5589,	56, 6,	and  992. The validation set is made up of 10\% of the training data. For experiments with pre-trained models, we use the checkpoints available at~\footnote{https://pytorch.org/vision/stable/models.html}.
\begin{table}[!ht]
    \centering
    \begin{tabular}{cc}
    \hline
    Hyper-parameter & Value\\
    \hline
       Learning rate  & $0.1$ \\
        Batch size & $128$ \\
        Weight decay & $5 \cdot 10^{-4}$ \\ 
        Training epoch & $200$\\
        $\alpha$ for LS & $0.1$ \\
        $\alpha$ for TF-reg & $0.1$ \\
        Temperature for TF-reg & $20$ \\
    \hline 
    \end{tabular}
    \caption{Hyper-parameters for ResNet18 on CIFAR100.}
    \label{tab:my_label}
\end{table}

\section{Hyper-parameter tuning}
For hyper-parameter tuning, we use the ray tune library ~\cite{liaw2018tune}. The tuned hyper-parameters are batch size, learning rate, $\alpha$, and temperature where they have been selected among \{8, 16, 32, 64\}, \{9e-6, 1e-5, 2e-5, 3e-5\},  \{0.4, 0.5, 0.7, 0.8, 0.9, 0.95\}, and \{1, 2, 5, 10\} sets respectively. We use ASHAScheduler algorithm of ray tune to find the best hyper-parameters. The metric of choosing them was maximum performance on dev set. The sample size of tuning hyper-parameters was 20 and 1 GPU was used for each experiment. Also, the maximum number of epochs for each trial was 20 epochs.

Since tuning hyper-parameters requires a huge amount of computational resources, we tried hyper-parameter tuning for vanilla KD on DistillRoBERTa model for GLUE benchmark. Then we chose the five best hyper-parameter sets from this experiment and checked their performance with other baselines and chose the one with highest average performance among all baselines. Only different $\alpha$ parameters for random initialization and pre-trained experiments on GLUE tasks and SST-5 made considerable differences in results. Therefore we used different $\alpha$ values in these experiments.

Tuning hyper-parameters individually for each baseline  would be a better option, but this would require a very large amount of computational resources. However, after careful hyper-parameter tuning for vanilla KD and less intensive hyper-parameter tuning for teacher-free baselines, the latter show very close performance to vanilla KD and this fact supports the main message of our paper.

\end{document}